\documentclass[journal, doublecolumn, 10pt]{IEEEtran}
\usepackage{setspace}
\usepackage{graphicx}
\usepackage{amsmath,amssymb,bm,amsfonts,amsthm}
\usepackage{color}
\usepackage{algorithm,algorithmicx}
\usepackage[noend]{algpseudocode}
\usepackage{multirow}
\usepackage{cite}
\usepackage{comment}

\renewenvironment{IEEEbiography}[1]
  {\IEEEbiographynophoto{#1}}
  {\endIEEEbiographynophoto}

\begin{document}

\title{Privacy-Preserving Serverless Edge Learning \\with Decentralized Small Data}
\author{ Shih-Chun~Lin and Chia-Hung Lin 
\IEEEcompsocitemizethanks{
\IEEEcompsocthanksitem Shih-Chun Lin and Chia-Hung Lin are with North Carolina State University.
\IEEEcompsocthanksitem This work was supported by Cisco Systems, Inc..
}
}
\markboth{Submitted for publication in the IEEE Network}%
\maketitle

\IEEEcompsoctitleabstractindextext{%
\begin{abstract} 
In the last decade, data-driven algorithms outperformed traditional optimization-based algorithms in many research areas, such as computer vision, natural language processing, etc. However, extensive data usages bring a new challenge or even threat to deep learning algorithms, i.e., privacy-preserving. Distributed training strategies have recently become a promising approach to ensure data privacy when training deep models. This paper extends conventional serverless platforms with serverless edge learning architectures and provides an efficient distributed training framework from the networking perspective. This framework dynamically orchestrates available resources among heterogeneous physical units to efficiently fulfill deep learning objectives. The design jointly considers learning task requests and underlying infrastructure heterogeneity, including last-mile transmissions, computation abilities of mobile devices, edge and cloud computing centers, and devices’ battery status. Furthermore, to significantly reduce distributed training overheads, small-scale data training is proposed by integrating with a general, simple data classifier. This low-load enhancement can seamlessly work with various distributed deep models to improve communications and computation efficiencies during the training phase. Finally, open challenges and future research directions encourage the research community to develop efficient distributed deep learning techniques.
\end{abstract}
}
\maketitle
\IEEEdisplaynotcompsoctitleabstractindextext
\IEEEpeerreviewmaketitle

\section{Introduction}\label{sec1}
Data-driven approaches have outperformed traditional optimization-driven algorithms in various research tasks recently. For optimization-driven methods, a mathematical system model should be provided in advance to reflect real-world phenomenons. However, the development of system models and the corresponding algorithms is non-trivial, hindering the advancement of optimization-driven algorithms. Instead, data-driven algorithms only need a training dataset. The relationship between input data and output labels can be automatically derived in the training process of deep learning algorithms, given pre-defined outputs (i.e., labels) to the training data. Hence, an efficient deep learning algorithm can still be developed, even without an accurate system model. 

Meanwhile, the extensive data usages bring a new challenge to deep learning algorithms, i.e., privacy-preserving. For example, in 2018, at least 87 million user information was disclosed in the Facebook Cambridge Analytica data scandal. Since then, governments around the world have been amending related laws to protect data collected from users. Also, as more than 90$\%$ of data will be stored and utilized in end devices, the traditional centralized training strategy becomes less attractive. 
First, as centralized training typically occurs in a cloud center environment, end devices need to upload local data to the cloud, generating compelling communications overheads to networking systems. Second, uploading local data to cloud centers is risky since the cloud operator will directly access user data. Last, the data transmitting process is fragile and vulnerable to malicious attacks. 
Hence, distributed deep training draws excellent attention. Each participant can keep sensitive data (e.g., medical profiles, financial profiles, and GPS historical data) in the local device to finish deep learning algorithms training. Communications overheads to upload local data to cloud centers can be avoided. 
However, the heterogeneity of end devices should be considered in the distributed training. Regarding the resource-limited nature of distributed training, how to improve the training efficiency by reducing the communication and computation overheads during the training phase becomes a vital research topic \cite{Commu-efficient1, Commu-efficient2, Compu-efficient1}. Also, incorporating other promising techniques (i.e., edge computing and AI-powered resource allocation) in distributed training scenarios to further improve training efficiency is another unsolved issue.

In this paper, we extend conventional serverless platforms \cite{serverless} with \textit{serverless edge learning architectures} and provide an efficient distributed training framework from the networking perspective. Notably, to better serve distributed training scenarios \cite{Our}, the proposed framework can dynamically allocate resources and intelligently assign communications and computation tasks to mobile devices, edge computation centers, and cloud computing centers. The intelligent allocation is based on the heterogeneous nature of wireless conditions, computation abilities, and battery status of devices. Most of existing privacy-preserving learning \cite{Privacy} can also be implemented in our framework and work with edge computing techniques to improve training efficiency. To further reduce distributed training overheads, we provide a learning enhancement that distributed training can be operated on small-scale data to increase communications and computation efficiency dramatically and simultaneously. While conventional communications-efficient or computation-efficient algorithms work on communications or computation efficiency separately, our design brings huge potentials to distributed training development with small data.

The rest of this paper is organized as follows. Section~\ref{sec2} reviews the latest progress of related research topics. Section~\ref{sec3} introduces distributed privacy-preserving learning over serverless edge architectures. Section~\ref{sec4} further provides a few-shot learning enhancement to realize small-scale data training. Section~\ref{sec5} lists open research directions, and Section~\ref{sec6} concludes the paper.

\section{State-of-the-Arts}\label{sec2}
In this section, we review the latest research progress of privacy-preserving learning and distributed training strategies.

\subsection{Privacy-Preserving Learning Approaches}
For privacy-preserving deep learning, training data and model parameters are the two elements required protection. In distributed training scenarios, data privacy is ensured as data is kept in local end devices and is not directly exposed to other participants. Hence, we focus on preserving model parameter privacy in distributed training \cite{Privacy}. First, several types of attacks can extract information from model parameters to reproduce training data. In a model inversion attack (i.e., reverse engineering attack), it is possible to reverse-engineer the model training by following the gradient during the model training to adjust the weights of an attacker model and obtain the original model's features. If model parameters are exposed during the training process, the training data is also revealed. Second, in applications of financial or high-tech industries, a sensitive model, which can perform accurately estimating (e.g., prices prediction or insurance rate estimation), can be considered as an essential commercial and intellectual property. In shadow training models, attackers can use the model parameters to train another model to achieve the same target, stealing confidential information to get benefits. 

To assure the privacy of model parameters, one can use \textit{federated learning} or \textit{differential privacy learning}. In particular, a typical working flow of distributed training can be divided into four steps\cite{DML}: (i) A central server broadcasts current model parameters to distributed workers (i.e., end devices). (ii) Workers use local data to train their models and obtain model updates. (iii) Each worker sends the model updates back to the central server. (iv) The central server aggregates received model updates from all workers to get updated model parameters. Such federated or differential privacy learning encrypts the downlink model parameters transmission in step (i) and the uplink model updates transmission in step (iii) to protect parameters exchanging. These privacy-preserving learning can be implemented in our proposed serverless edge architecture as upper applications.

\subsection{Distributed Deep Training Strategies}
The goal is to maintain the achieved performance while reducing communications or computation overheads. To improve communications efficiency \cite{Commu-efficient1, Commu-efficient2}, coding or model compression schemes (e.g., model pruning, model quantization) can be applied on model parameter transmissions in step (i) and model update transmissions in step (iii). However, these schemes work after local model training and cannot improve the computation efficiency, and they also create computation burden by performing additional compression schemes.

On the other hand, to improve computation efficiency \cite{Compu-efficient1}, existing algorithms introduce the concept of importance sampling or similarity calculation (i.e., active sampling) to choose the samples with higher importance or lower similarity for the model training to mitigate computation overhead. However, these approaches also bring a new challenge to distributed training, as estimating the behavior of a neural network on a specific sample is very difficult. Computation-efficient algorithms need remarkable computations, which cannot be ignored in the distributed training scenarios, to perform accurate estimations, threatening their practicality.
In short, the development of distributed deep training strategies is still in its infancy stage and requires innovations. 

\section{Efficient Distributed Privacy-Preserving Learning over Serverless Edge Learners}\label{sec3}
We introduce a serverless edge learning framework to facilitate distributed training in modern networking systems. 

\subsection{Serverless Edge Architecture}
With the serverless edge learning framework, communication and computing resources in the whole networking system can be managed and leveraged to accelerate deep learning algorithms' training in distributed scenarios even with the heterogeneity of end devices. Privacy-preserving learning algorithms can also be executed efficiently to protect model parameters during distributed training with efficiency, especially in the case of additional computations required for those privacy-preserving algorithms to perform secure data exchanging. The achieved benefits of serverless edge learning framework can be listed below: (i) As for application developers, developers do not need to manage the resource allocation or handle the burden of scalability. Instead, a central controller of the serverless edge learning framework will manage the networking system to facilitate the distributed training process. (ii) Users only need to submit requests to serverless edge learning framework to perform learning tasks. The serverless edge learning framework will allocate communication and computation resources to satisfy the requested learning tasks. (iii) From a networking system perspective, underutilization of limited edge resources can be prevented due to the centralized resource allocations in the serverless edge learning framework.   

As shown in Fig. 1, the whole framework can be divided into three parts: clients, serverless edge computing network, and cloud centers in the serverless edge learning structure. The clients refer to end devices such as mobile devices or laptops, which are with data in devices and aim to perform learning tasks using local data. The serverless edge computing network includes a controller responsible for global network control and resource orchestration for the serverless edge learning framework and several distributed serverless edge computing nodes. Those edge computing nodes can be regional cloud or edge cloud computing centers according to the topology of the interested networking system. Last, the cloud centers refer to powerful computing centers, which are usually far from clients geographically. 
\begin{figure*}
    \centering
    \includegraphics[width=0.85\linewidth]{ 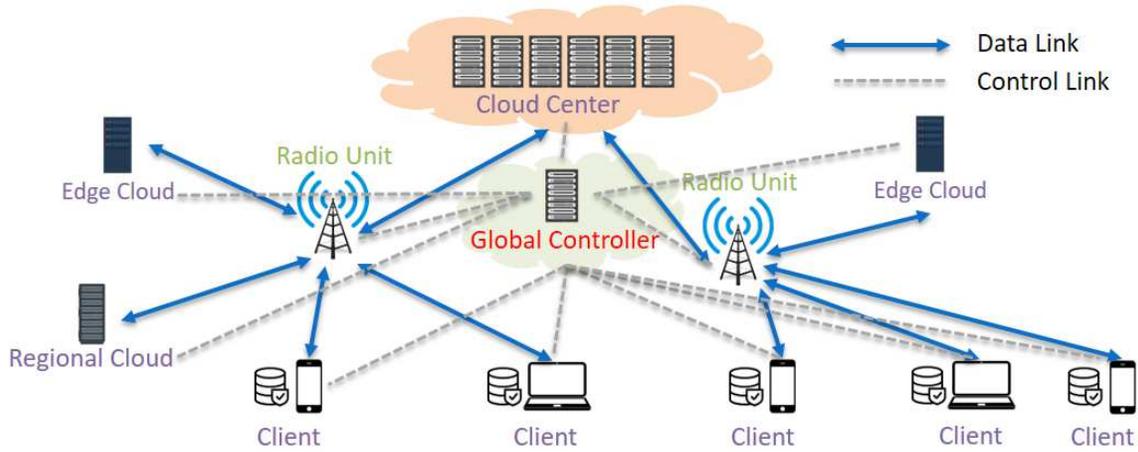}
    \caption{The consider scenario of serverless edge learning framework in a distributed environment. There are three parts of objects in the serverless edge learning framework: clients, edge computing nodes, and cloud computing centers. All the objects need to report resource and request status to the global controller. This controller will perform learning tasks allocation based on resource and request status, realizing global management and distributed computing.}
    \label{fig:Sys}
\end{figure*}

When a requested training task belonging to latency-sensitive applications (such as autonomous vehicles and augmented reality) arrives at serverless edge architectures, the controller can allocate the learning task to nearby edge cloud computing centers to achieve minimum latency via its centralized resource allocation capability. On the other hand, for a computational-demanding training task, such as BigGAN \cite{BigGAN} and large-scale text generation, the controller intends to assign the related computations to the cloud centers to enjoy the powerful computation abilities. To realize the aforementioned \textit{global control and distributed computing}, all the clients, edge computing nodes, and cloud centers are required to report the communication and computation resource status to the controller. Simultaneously, all the clients also need to send requests to the controller to train learning tasks in a serverless edge learning structure. Two main modules should be implemented on the global controller to perform the information collections effectively and efficiently, as shown in Fig. 2, the service management module and network and resource management module. The network and resource management module is responsible for collecting global network information and resource information, and the service management module focuses on the collection of learning service requests. After receiving all the resources and request information, the designed controller can regulate the intelligent networking infrastructure to manage all resources, including computation and communication, according to the quality of service (QoS) demands of clients' learning services.

Based on the serverless edge learning structure, we implement a serverless edge learning framework using Python and Tensorflow \cite{TF} to perform distributed training and inferring. Specifically, we use different computational components (i.e., CPU, GPU, TPU) in a single server to represent heterogeneous objects in serverless edge learning structure and employ Google remote procedure calls (gRPC) \cite{gRPC} for data transmission between objects to finish the training of deep learning algorithms in a distributed manner. The developed framework can be further extended to different computers in a cluster with an internet connection to perform deep learning training experiments with geographically non-adjacent computers. As for the global controller, an application interface (API) that clients can use to describe the desired QoS is provided in terms of latency, bandwidth, priority, and computing resources. Also, another API is offered to computation nodes for reporting resource status to the global controller. Therefore, the global controller can monitor the utilization and capabilities of resources in the whole network, including edge nodes, cloud services, and end devices. More important, this controller can dynamically adjust the workloads among different computing resources across the whole network, considering the real-time capability of each component with the help of intelligence within networking nodes (e.g. routers with computing capabilities). Finally, our framework can further use machine learning models to adaptively assist those decisions if the infrastructure contains free resources to perform these models' necessary computation. The global controller is used to perform computation resource allocations considering available computation resources in networking systems to satisfy learning requests from clients and additional calculations from performing privacy-preserving learning.
\begin{figure*}
    \centering
    \includegraphics[width=0.85\linewidth]{ 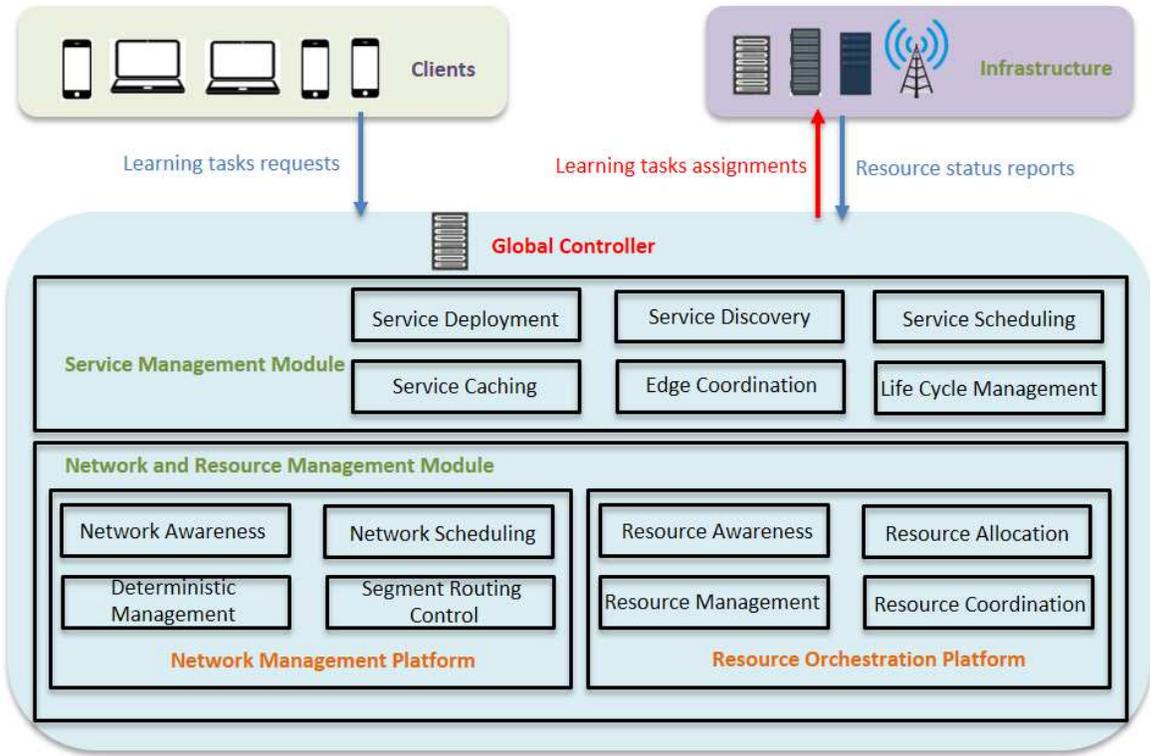}
    \caption{The function blocks of the serverless edge learning framework. With the provided resource status and learning tasks requests, the global controller is responsible for performing service management and network and resource management to facilitate distributed deep learning training efficiently.}
    \label{fig:Sys}
\end{figure*}

\subsection{Distributed Privacy-Preserving Learning}
With the developed platform, we can evaluate different training strategies in a distributed environment. For example, a well-known MNIST handwriting image classification task is considered and implemented in a distributed scenario to obtain simulation results. We have four clients. Each client holds non-overlapping data and a local deep learning model to learn image classification tasks. In each epoch, each client will update its local model using its own local data independently. After that, a client will be responsible for collecting and aggregating all updated models, then broadcasting the aggregated results back to all clients to train the next epoch. We present two simulation results to show the training results with different distributed scenarios. First, we are curious about how the data distribution among clients affects the achieved performance results of deep learning algorithms. We assume the computation abilities of each end device remain the same during the whole training period (i.e., all end devices are the same mobile devices). In an MNIST training dataset with $55000$ images, we set the training data in each client as $[25000, 25000, 4000, 1000]$. Each client can only process 1000 training samples in each epoch due to the computing power limitation. Here, we present the learning curve of two cases. First, all end devices choose 1000 training data and use those data to finish the training of all epochs. Second, at the beginning of each epoch, all end devices can randomly choose 1000 training data again to complete the training of all epochs.

The results are shown in Fig. 3. Although the number of training data is limited, the learning curves show that convergence can still be achieved in both cases. Also, the learning curves reveal that simple random sampling can be used to maintain performance in distributed deep learning scenarios effectively. More complicated training data sampling schemes may not be required in a distributed training of MNIST classification task. Second, we aim to discuss how the unequal updating frequencies of each client affect the achieved performance of the deep learning model in the distributed deep learning training scenario. This effect can be caused as a result of unequal computation abilities of each client. To be more specific, this phenomenon is so-called gradient staleness \cite{staleness} in academia. To explain, gradient-based optimization algorithms, which are usually employed in the deep learning training process, belong to iterative algorithms. In each iteration, we need to set current trainable weights as a starting point and compute gradient information to update trainable weights based on the starting point. If each client's unequal updating frequencies occur, all the gradient information provided by each client does not start from the same starting point. As a result, the aggregation results of gradient information may no longer point to optimal optimization direction for trainable weights updating. However, to the best of our knowledge, how the gradient staleness issues affect the achieved performance of different applications is not well investigated. 

Moreover, even when the gradient staleness issues affect performance severely, existing algorithms to deal with gradient staleness issues only use a passive way by adjusting the weights of gradient staleness to maintain the aggregation results. This decision does not strike a good trade-off between the computational abilities of each client and achieved performance. In this simulation, with equal data distribution among each client, we set the updating frequencies of each client as $[20, 20, 1, 1]$ (i.e., Unequal in Fig. 3) and compare the results with a standard case, which is $[20, 20, 20, 20]$ (i.e., Equal in Fig. 4) for updating frequencies of each client. In each epoch, the first two clients in the unequal case can finish $20$ iterations of gradient descent algorithm, while the last two clients in the unequal case can only finish one iteration.  As shown in Fig. 4, it is clear that the gradient staleness issue results in slower convergence in distributed deep learning training, revealing that novel distributed deep learning algorithms should be developed to tackle possible gradient staleness issues.
\begin{figure}[!t]\centering
\includegraphics[width=3.4 in]{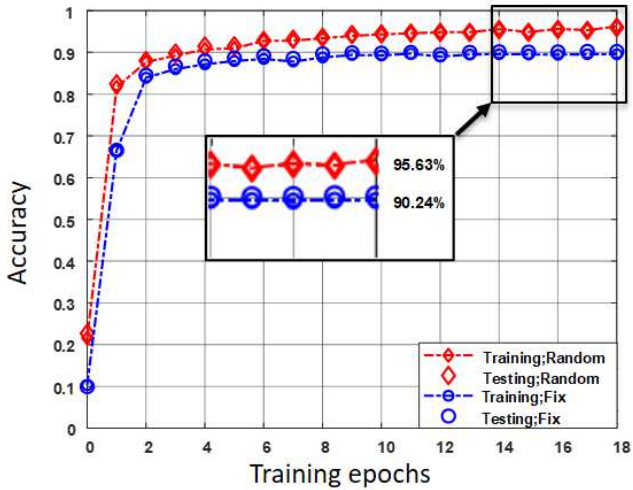}
\caption{The achieved performance of different sampling strategies under unequal data distribution among clients in a distributed deep learning training scenario.}\label{MACC1}
\end{figure}

\begin{figure}[!t]\centering
\includegraphics[width=3.4 in]{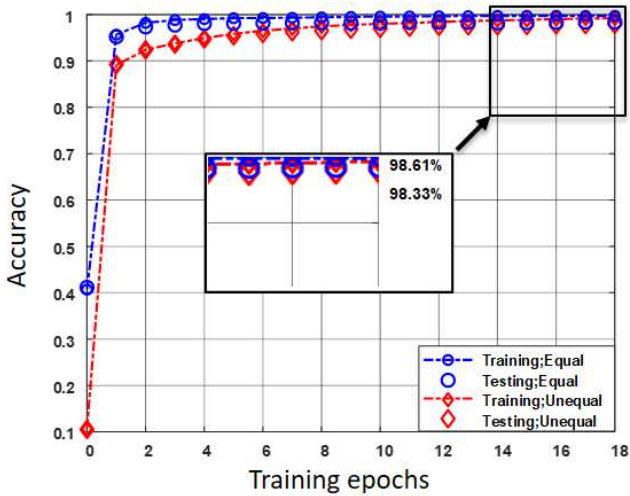}
\caption{The achieved performance of different gradient staleness statuses among clients in a distributed deep learning training scenario.}\label{MACC1}
\end{figure}

\section{Few-Shot Learning Enhancement}\label{sec4} 
We provide a novel enhancement for efficient distributed training in Section~\ref{sec2} and study a use case to show that communications and computation overheads during distributed training can be improved simultaneously. Thus, this enhancement can bring a better trade-off between achieved performance and training overheads in serverless edge learning. 

\subsection{Prototype Network Implementation}
Users are motivated to employ deep learning models with more trainable parameters to increase the capacity of deep learning models for better performance, adding communication and computation overheads during training at the same time. Alternatively, novel few-shot learning \cite{FS1} matches the need for distributed deep training strategies naturally. To be more specific, few-shot learning aims to classify new data, having seen only a few training samples. Unlike conventional deep learning, few-shot learning algorithms combine classical data-driven schemes (e.g., $k$-nearest neighbor algorithm) and deep learning models to limit the number of trainable parameters as those classic data-driven schemes can be used to reach the same goal with fewer or even no trainable parameters. As the number of trainable parameters in few-shot algorithms is reduced significantly, the number of training data can also be reduced as long as the overfitting does not occur.

Based on this concept, we aim to build a specialized neural network containing a lower amount of trainable parameters to address the training challenges of distributed deep learning. As a result, the computation and communication efficiency can be improved simultaneously as the number of training samples and parameters are reduced. However, there are no such studies in the literature to investigate the benefits of employing few-shot learning to aid the distributed deep training. Therefore, we present a case study here to show the potential of this research direction and encourage researchers to contribute efforts to it. Toward this end, we develop our solution based on a prototype network \cite{Prototype}, a classic algorithm of few-shot learning, and evaluate the benefits in distributed deep learning scenarios. The idea of the prototype network is to employ a deep learning model to perform feature extraction automatically, then use the concept of $k$-nearest neighbor to finish the classification task. Specifically, the deep learning model is responsible for performing feature extraction to project the original input sample to a corresponding feature vector. It is good at extracting useful high-level features to aid the decision process. Then, a traditional classifier is adopted with significantly lower or even no trainable parameters to generate classification results.

Given training samples belonging to each class, the deep learning-based feature extraction module will be used to extract high-level features from the training samples to present corresponding feature vectors in feature space with a fixed dimension. For each specific class, the mean value of that class (i.e., the prototype) in the feature space can be calculated by computing the arithmetic average of feature vectors belonging to the case. When a new data sample is obtained for classification, the deep learning-based feature extraction module will be employed again to project the data point to the feature space. Then the probability that the data point belongs to each class can be calculated by the distance between the feature vector of the new sample and the mean values of training samples belonging to each class. The whole process is also illustrated in Fig. 5. In order to minimize the cross-entropy loss function of the prototype network, the deep learning-based feature extraction module is forced to find the best projection way to present data in the feature space, consequently improving the final classification accuracy.
\begin{figure*}[!t]
\centering
\includegraphics[width=0.85\linewidth]{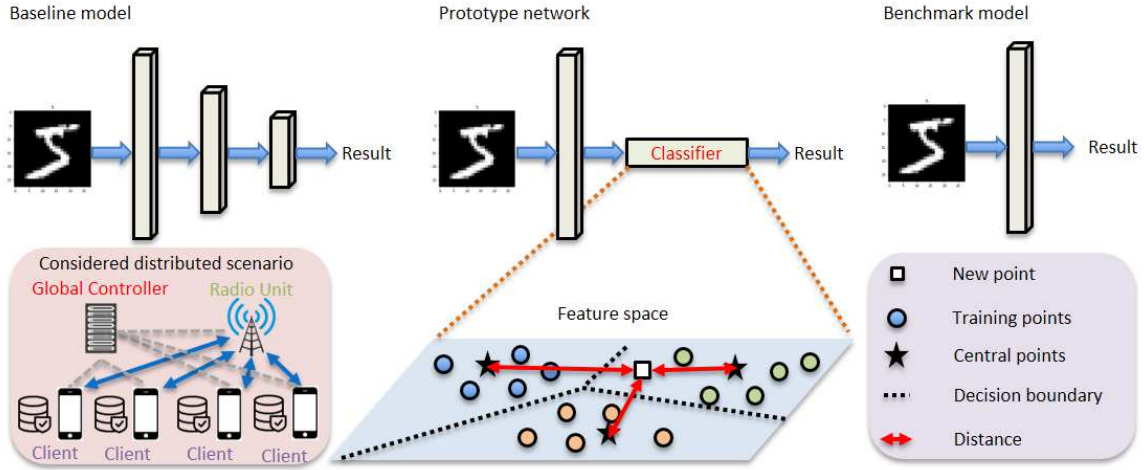}
\caption{The model architectures of baseline, prototype, and benchmark models and the classification mechanism of the prototype network. In particular, we implement these models in a considered distributed scenario using a serverless edge learning framework to evaluate the training overheads of different models.}
\label{Prototype}
\end{figure*}

\subsection{Performance Evaluation of Prototype Network}
We implement the prototype network in a serverless edge learning framework and compare the results with two competitors \cite{MNIST} to show the potential of few-shot learning algorithms in distributed deep learning training environment.  As in the previous evaluation, the MNIST handwriting image classification task is implemented in distributed deep learning scenario to obtain the simulation results in this section. We also set the number of clients as four, and each client holds non-overlapping data and a local deep learning model to perform image classification.

To show the related communication and computation overheads of different algorithms, we define the total number of trainable parameters, which should be transmitted between clients to finish the model training, as communication overhead. On the other hand, the total number of needed training samples, which should be proceeded in each client to finish the model training, is defined as computation overhead. The ratios between achieved overheads of different algorithms and the overheads of a baseline model are considered performance indicators. As shown in Fig. 5, we set a deep learning model with three layers as the baseline model for comparison \cite{MNIST}.  As for the prototype-based deep learning model, we construct a feature extractor with a single-layer neural network and employ a $k$-nearest neighbor algorithm-based classifier to finish the classification task. As the number of trainable parameters of the prototype-based model decreases, the required training data can be reduced. We also build a benchmark model \cite{MNIST}, which is a single-layer neural network model with the same communication and computation overheads as the prototype-based deep learning model.

The results in Table I indicate that with the aid of the prototype network, the computation and computation efficiency improve significantly. Only 1.81$\%$ of computation and 4.43$\%$ of communication resource should be employed in the prototype network in a distributed deep learning scenario. On the other hand, without a prototype network, the accuracy of the single layer model presents a 15$\%$ loss with the same communication and computation overheads, reflecting the potential to develop prototype-based distributed deep training. Notably, the prototype network almost does not increase any training overheads in distributed deep learning scenarios compared to the benchmark model. Also, our scheme can be considered as a specially-designed model architecture of distributed deep learning algorithm. Hence, our scheme can work with most existing communication-efficient, computation-efficient, and optimizers for distributed deep learning scenarios to enjoy the provided benefits if needed.
\begin{table}\caption {Performance and overhead comparisons among different distributed training schemes.}
\begin{tabular}{|c|c|c|c|}
\hline
                                                          & \begin{tabular}[c]{@{}c@{}}Achieved\\ accuracy\end{tabular} & \begin{tabular}[c]{@{}c@{}}Communications\\ overhead\end{tabular} & \begin{tabular}[c]{@{}c@{}}Computation\\ overhead\end{tabular} \\ \hline
\begin{tabular}[c]{@{}c@{}}Baseline\\ model\end{tabular}  & 97.98\%                                                     & 100\%                                                            & 100\%                                                          \\ \hline
\begin{tabular}[c]{@{}c@{}}Prototype\\ model\end{tabular} & 87.74\%                                                     & 4.43\%                                                           & 1.81\%                                                         \\ \hline
\begin{tabular}[c]{@{}c@{}}Benchmark\\ model\end{tabular} & 73.24\%                                                     & 4.43\%                                                           & 1.81\%                                                         \\ \hline
\end{tabular}
\end{table}

\section{Challenges and Research Directions}\label{sec5}
In this section, we present open challenges for performing efficient distributed training by serverless edge learners.

\subsection{Heterogeneity of Distributed Learning}
The heterogeneity is the most challenging part of distributed deep training in serverless edge architecture. Computation and communication resources of end devices, edge computation nodes, and cloud centers are different. Also, all objects' computation and communication abilities will even become time-varying when we consider wireless links between nodes and the battery status of end devices. As a result, the design of resource orchestration modules is challenging to satisfy different QoS requirements. Intelligent resource management should be developed to dynamically adjust the workloads among various computing resources across the entire network while considering the real-time capability of each component. Deep reinforcement learning algorithms can be good candidates to meet the need. Specifically, reinforcement learning is suitable to design an adaptive resource association to guarantee QoS over time. Deep learning schemes \cite{ResourcePrediction} can further predict future resource usages; thus, reducing response time for serverless edge infrastructures.    

\subsection{Resource-Demanding of Deep Learning Algorithms}
The resource-demanding nature is another severe challenge to distributed deep learning training in serverless edge learning framework. Some deep learning algorithms with more than a billion trainable parameters \cite{BigGAN}  can achieve good performance. They also require a large-scale training dataset, further increasing the computation and communications burdens. Hence, training such algorithms in a distributed scenario is pretty challenging. The proposed few-shot learning algorithms can address this issue by designing specialized model architectures to distribute these heavy-loaded deep algorithms. Notably, the combination of simple classical classifiers and deep learning algorithms can indeed strike a better balance between achieved performance and needed training overheads. 

In addition, transfer learning and neural network architecture are two open research directions in developing efficient deep learning. First, essential features for learning development are mostly task-independent. While focused tasks are different in different research areas, basic learning features might be highly relevant. This similarity suggests transfer learning algorithms to perform \textit{intelligent caching}, facilitating deep training in a distributed manner. Remarkably, our serverless edge architectures can extract and store knowledge with transfer learning. Then, when a client wishes to train a new deep learning algorithm, the accumulated knowledge can accelerate the convergence, decreasing needed communications and computation resources. Second, fully-connected neural network architecture often introduces a considerable number of trainable parameters. As an alternative, some recent studies show better efficiency with graph neural networks and transformer neural networks. Adopting these two kinds of architectures can significantly reduce the number of trainable parameters, easing communications and computation burdens in distributed deep learning cases.

\subsection{Privacy-Preserving of Distributed Deep Training}
Additional computation tasks generated from privacy-preserving learning is also a severe challenge to serverless edge learning framework. Specifically, to safely exchange model parameters in serverless edge learning structures, federated or differential privacy learning should be employed to perform secure model parameter transmission. However, existing schemes still introduce a significant computation burden to design encryption and decryption mechanisms for privacy preservation. In this direction, low complexity encryption and decryption mechanisms should be further investigated to release the computation burden to preserve privacy. Blockchain-based encryption can also be considered to transmit model parameters with low complexity securely.

\section{Conclusion}\label{sec6}
We propose a serverless edge learning framework to aid distributed training efficiency from the networking perspective. The framework achieves dynamic orchestration to utilize networking system resources considering learning task requests and underlying infrastructure heterogeneity. We also demonstrate the potential of few-shot learning in distributed training scenarios by evaluating the reduced communication and computation overheads during the training phase. Open challenges and future research directions of serverless edge learning are summarized to encourage the development of efficient distributed training strategies.



\bibliography{MLNSmag}
\bibliographystyle{IEEEtran}

\end{document}